\documentclass[10pt,twocolumn,letterpaper]{article}

\usepackage{wacv}              % To produce the CAMERA-READY version
\usepackage{graphicx}
\usepackage{amsmath}
\usepackage{amssymb}
\usepackage{booktabs}

\usepackage{pifont}% http://ctan.org/pkg/pifont

\usepackage[pagebackref,breaklinks,colorlinks]{hyperref}

\usepackage{subcaption}
\usepackage{booktabs}
\usepackage{multirow}
\usepackage{xcolor,colortbl}
\definecolor{MyPink}{RGB}{255, 221, 219}
\definecolor{MyBlue}{RGB}{0, 0, 255}

\usepackage[capitalize]{cleveref}
\crefname{section}{Sec.}{Secs.}
\Crefname{section}{Section}{Sections}
\Crefname{table}{Table}{Tables}
\crefname{table}{Tab.}{Tabs.}

\def\wacvPaperID{1697} % *** Enter the WACV Paper ID here
\def\confName{WACV}
\def\confYear{2024}

\begin{document}

\title{FLORA: Fine-grained Low-Rank Architecture Search for Vision Transformer}

\author{
\hspace{-2em} Chi-Chih Chang$^{1,*}$, Yuan-Yao Sung$^{1,*}$, Shixing Yu$^{2,3,*}$, Ning-Chi Huang$^{1}$, Diana Marculescu$^{2}$, Kai-Chiang Wu$^{1}$\\
\hspace{-2.5em} National Yang Ming Chiao Tung University$^{1}$, University of Texas at Austin$^{2}$, Cornell University$^{3}$ \\
{\tt\small brian1009.en08@nycu.edu.tw, sungyuanyao@gmail.com,} \\
{\tt\small sy774@cornell.edu, dianam@utexas.edu, \{nchuang, kcw\}@cs.nctu.edu.tw}
}

\maketitle
\begin{abstract}
Vision Transformers (ViT) have recently demonstrated success across a myriad of computer vision tasks. However, their elevated computational demands pose significant challenges for real-world deployment. While low-rank approximation stands out as a renowned method to reduce computational loads, efficiently automating the target rank selection in ViT remains a challenge. Drawing from the notable similarity and alignment between the processes of rank selection and One-Shot NAS, we introduce FLORA, an end-to-end automatic framework based on NAS. To overcome the design challenge of supernet posed by vast search space, FLORA employs a low-rank aware candidate filtering strategy. This method adeptly identifies and eliminates underperforming candidates, effectively alleviating potential undertraining and interference among subnetworks. To further enhance the quality of low-rank supernets, we design a low-rank specific training paradigm. First, we propose weight inheritance to construct supernet and enable gradient sharing among low-rank modules. Secondly, we adopt low-rank aware sampling to strategically allocate training resources, taking into account inherited information from pre-trained models. Empirical results underscore FLORA's efficacy. With our method, a more fine-grained rank configuration can be generated automatically and yield up to 33\% extra FLOPs reduction compared to a simple uniform configuration.
More specific, FLORA-DeiT-B/FLORA-Swin-B can save up to 55\%/42\% FLOPs almost without performance degradtion. Importantly, FLORA boasts both versatility and orthogonality, offering an extra 21\%-26\% FLOPs reduction when integrated with leading compression techniques or compact hybrid structures. Our code is publicly available at \mbox{\url{https://github.com/shadowpa0327/FLORA}}.
%\href{https://github.com/shadowpa0327/FLORA}

\end{abstract}
\vspace{-10pt}
\section{Introduction}
\label{sec:intro}

The transformer architecture \cite{DBLP:conf/nips/VaswaniSPUJGKP17} has dominated natural language processing (NLP) tasks with impressive results.
Though intuitively, the transformer model seems inept to the special inductive bias of space correlation for image-oriented tasks, it has proved its capability on vision tasks with comparable results to convolutional neural network (CNN)~\cite{DBLP:conf/iclr/DosovitskiyB0WZ21}. Since their inception, vision transformers (ViT) and their variants have shown great potential for image classification \cite{DBLP:journals/corr/abs-2106-04560}, object detection \cite{DBLP:journals/corr/abs-2205-14141}, and semantic segmentation \cite{DBLP:conf/iccv/LiuL00W0LG21}.
However, the ViT requires a large number of parameters and high computational cost to obtain higher accuracy, making it unsuitable for edge computing. That is mainly due to the stack of self-attention modules that suffer from quadratic complexity with regard to the input size, among other factors.
Hence, research on efficient transformer models has become more important recently.

Earlier works on compressing ViTs mainly follow the techniques for compressing NLP models, ranging from unstructured pruning \cite{zhu2021visual}, attention head/structured pruning~\cite{DBLP:conf/iclr/YuCSYTY0W22, DBLP:conf/nips/ChenCGYZW21}, token pruning~\cite{DBLP:journals/corr/abs-2112-13890, DBLP:conf/nips/RaoZLLZH21, yin2022vit}; to knowledge distillation \cite{DBLP:conf/icml/TouvronCDMSJ21,jia2021efficient} and quantization \cite{yuan2021ptq4vit, liu2021post}.

Aside from the above-mentioned directions, another important category of method that employ efficiency in neural network (NN) structure is low-rank approximation. %These methods have already proven its success in convolution neural network (CNN)~\citep{DBLP:conf/cvpr/YuLWT17, DBLP:journals/tcas/ChenCLLL20}.
In the case of 2D low-rank approximation, singular-value decomposition (SVD) minimizes the Frobenius norm of the difference between the original matrix and the approximated matrix. Yet, SVD cannot be directly utilized for convolutions in CNNs because weights need to be represented by higher-dimensional (\emph{e.g.,} 4D) tensors~\cite{DBLP:journals/corr/LRACNN}. Special design~\cite{DBLP:journals/corr/KimPYCYS15, DBLP:journals/corr/LebedevGROL14} is developed for CNNs by decomposing them into multiple consecutive tensors. However, there is still significant accuracy drop even after fine-tuning as training is performed on the transformed network structure with consecutive tensors without activation functions in-between. As a result, convergence can be degraded due to vanishing or exploding gradients~\cite{DBLP:journals/corr/LRACNN}. 
This obstacle is largely alleviated in ViTs since 90\% of total parameters and operations (see the supplemental materials for detailed analysis) are linear modules and conduct matrix multiplication. Linear modules can be decomposed into just two consecutive matrices. 
To put it in another way, ViTs are much more friendly than CNNs when incorporating low-rank decomposition in their structure.
Meanwhile, low-rank decomposition considers redundancy in deep neural networks as noise that contains a very small percentage of variance. Hence, it's different from general pruning methodologies for NN compression, in the way that the overall structural dimension and information flow will not be affected or truncated through low-rank guided compression techniques.

To effectively integrate low-rank approximations without compromising network performance, the central challenge lies in identifying a suitable set of rank settings. In the context of a Vision Transformer (ViT) with $N$ linear modules and a maximum rank value of $M$, the potential rank settings explode to an immense number of $M^{N}$. Typically, $M$ takes values in the order of a few hundred, leading to an exceedingly vast array of choices. However, existing methods that deal with low-rank approximation for transformers often resort to heuristic rank choices\cite{DBLP:conf/acl/LvZLGS23} or enforce uniform ranks across all compression targets\cite{DBLP:conf/icassp/WinataCLLF20}, due to the absence of an automated rank selection strategy. Clearly, there is a strong need for an automatic rank selection approach to reduce the resource-intensive search process and mitigate the potential for suboptimal outcomes.

%To surmount this challenge and encourage the widespread adoption of low-rank approximation, 
To surmount this challenge we present an innovative approach: reimagining the rank selection process as a Neural Architecture Search (NAS) problem. This new viewpoint is based on two main insights.

First, finding the best rank is a lot like choosing the right architecture for low-rank models. This similarity arises because picking the rank essentially   how the corresponding low-rank architecture is built underneath.

Second, a property of SVD-based low-rank architectures, specifically the inheritance of top-$r$ eigenvectors from pretrained. For different rank settings $r_i$ and $r_j$, where $r_i < r_j$ the corresponding low-rank architecture share a significant portion of information from pretrained weights at the initial state. Such intrinsic property offers a great opportunity for us to train these candidates jointly like the way the supernet is optimized via weight sharing strategy in One-Shot NAS\cite{DBLP:conf/pkdd/StamoulisDWLPLM19, DBLP:conf/eccv/GuoZMHLWS20}. Building upon the well-recognized principles of One-Shot NAS, we introduce Fine-grained Low-Rank Architecture Search, dubbed as FLORA, to search for optimal low-rank architecture for ViT. Our main contributions are outlined as follows:
\begin{itemize}

\item We have discovered a captivating correlation between rank selection and One-Shot NAS and propose a first-of-its-kind low-rank architecture search regime for vision transformer.

\item To realize our vision, we propose a low-rank aware filtering policy specifically crafted to eliminate candidates exhibiting subpar performance. By mitigating interference from less promising alternatives, this approach directly addresses the training challenges a supernet encounters due to the vast low-rank search space, thereby enhancing the efficiency and effectiveness of the overall NAS process.

\item Building on this, we architect a unique supernet training paradigm tailored for low-rank structures, leveraging the inherent properties of SVD. This promotes both parameter and gradient sharing, significantly accelerating supernet convergence. As a result, subnetworks emerge with competitive performance, primed for direct deployment without exhaustive retraining.

\item We conduct a series of experiments to demonstrate FLORA's orthogonality to other compression techniques and its generalization capability on ViT variants, which provides a new perspective on high ratio compression for ViT.
% , including the hybrid structure (CNN + Transformer). 
Extensive experiments further demonstrate competitive results on ImageNet-1k when compressing DeiT and Swin-Transformer. For instance, FLORA-DeiT-B and FLORA-Swin-B achieve FLOPs reductions of 55\% and 46\%, respectively, on their corresponding backbones with negligible accuracy degradation.
%Benefited by superior teacher, we can further improve our 
%We also make... 
\end{itemize}

\section{Related Work}
\label{sec:related}

\subsection{Transformer Compression}
\label{Related_work:Pruning}

%This property is inherited to Vision Transformer is used to tackle vision tasks
%While transformers have the flexibility to scale to high parametric complexity, they impose a heavy computational burden, which includes huge training and inference costs. This indicates the necessity for efficient vision transformer models. In this section, we discuss some methods that carried out model compression and efficient transformer design.

%In transformer-based models, multiple attention operations perform in parallel\cite{DBLP:conf/naacl/DevlinCLT19, DBLP:conf/nips/VaswaniSPUJGKP17}. However, \cite{DBLP:conf/emnlp/KovalevaRRR19, DBLP:conf/nips/MichelLN19, DBLP:conf/acl/VoitaTMST19} demonstrated that numerous heads in the multi-head attention module are redundant and can be removed without significantly degrading model performance. \cite{DBLP:conf/iclr/FanGJ20, DBLP:conf/nips/HouHSJCL20} showed the less important layers in a transformer model can be pruned to obtain a shallower model. VTP\cite{zhu2021vision} reduced the number of embedding dimensions by extending the network slimming approach \cite{DBLP:conf/iccv/LiuLSHYZ17} to vision transformers. Some pruning algorithms have been proposed for vision transformers that split an image into multiple patches. For instance, \cite{DBLP:journals/corr/abs-2106-02852, DBLP:conf/nips/RaoZLLZH21} designed patch pruning based methods to discard useless patches and reduce patch calculation.\par
Compression methods for transformers can be broadly categorized into: pruning, token reduction and efficient architecture design. 
Pruning techniques are proposed to alleviate the high computational cost and memory usage by removing the redundant weights in the transformer-based models.
VTP \cite{zhu2021visual} reduced the number of embedding dimensions by extending the network slimming approach \cite{DBLP:conf/iccv/LiuLSHYZ17} to ViTs.
\cite{DBLP:conf/iclr/FanGJ20, DBLP:conf/nips/HouHSJCL20} proposed to skip the inessential layers to obtain a shallow model.
Similarly, WDPruning \cite{DBLP:conf/aaai/0004HWCCC22} removed the less significant channels of the linear projection by using a neural-network-based saliency predictor.
 %\cite{DBLP:conf/aaai/0004HWCCC22} 
For the token-reduction-based techniques, \cite{DBLP:journals/corr/abs-2106-02852, DBLP:conf/nips/RaoZLLZH21} hierarchically remove the redundant patches, thus reducing the computational overhead by slimming the input features.  

Aside from above, some other works dedicated on design a efficient architecture directly by introducing CNNs to form a hybrid structure. MobileViT  \cite{DBLP:conf/iclr/MehtaR22} mixed global processing in transformers with convolution, which learns better representations with fewer parameters and simple training recipes. 

Recent work endeavored to combine multiple pruning strategies into a unified framework, \emph{i.e.}, considering multiple dimensions simultaneously. For instance, UVC \cite{DBLP:conf/iclr/YuCSYTY0W22} pruned the heads in MHSA, channels in linear projection, and considered layer skipping. Meanwhile, MDC \cite{DBLP:conf/cvpr/HouK22a} jointly optimized an extra dimension, the number of patches.

\subsection{Low-Rank Approximation}
%\chichih{Hi, shixing: First, thanks for you nice suggestion. We have add the related work of low-rank approximation. Could you please read it again and give us some feedback?}
%\shixing{Hi chichih, sorry that I missed some key points of the methodology in this section due to some misunderstanding. I thought we have also decomposed the self-attention calculation also with low-rank, it turns out that we just did it in the projection module right? If that's the case, I would suggest that we only introduce Low-rank Approximation in general, instead of Low-rank for self-attention, sorry for the misunderstanding.}

% \chichih{Hi, shixing, some of the low-rank related work we have surveyed was add at the google docs, you may take a look, if you need it.}
% \shixing{I see. Thanks!}
% \shixing{Hi Chichih, is there something lost before this part? Since it starts from "Aside from the prior arts mentioned above"}

% \chichih{No, the prior arts mentioned above means the model compression paradigm we have summarized in previous section. Maybe we can mention it explicitly, to avoid confusing? }
% \chichih{Something like, "Aside from the model compression technique mentioned above" ?}
% \shixing{Yes, this makes better sense.}
% \chichih{Got it, I have replaced the original sentence into the new one.}
Aside from the model compression technique mentioned in the previous section, the low-rank matrix factorization on weights is also an effective methodology to reduce computational burden. In CNN-based models, \cite{DBLP:journals/corr/TaiXWE15} split the convolution kernel into two small ones with fixed rank. Some other works studied rank selection to generate finer-grained low-rank approximation. \cite{DBLP:conf/ijcai/XuL0WWQCLX20} selected the rank base on thresholding. \cite{DBLP:conf/cvpr/IdelbayevC20} introduced rank-based cost function and formulated a constraint optimization problem to decide the rank selection during training. 

Low-rank approximation (LRA) for transformer-based models can mainly be divided into two categories: (1) LRA
for attention matrix (ScatterBrain\cite{DBLP:conf/nips/ChenDWSRR21}, OmniNet\cite{DBLP:conf/icml/Tay0AGP0BJM21}) and (2) LRA for linear embedding (LRT \cite{DBLP:conf/icassp/WinataCLLF20}). The former is mainly designed for transformers in NLP whose input sequences are relatively long. When the sequence length (patch number) is small (\textit{e.g.}, ViT), the performance gain is relatively minor, as shown in the original paper of ScatterBrain. As for the latter one, LRT applies LRA with uniform-rank config and demonstrates its potential on Transformer for speech recognition tasks. Besides, we empirically found that the sensitivity to the performance loss concerning rank level is different among each linear embedding layer. Therefore, in this work, we set up a novel paradigm based on One-Shot NAS to explore the immense rank selection space of ViT more thoroughly and reduce the potential sub-optimality incurred by the simple manual setting.

%However, due to the vast search space of low-rank the effect of low-rank approximation on linear layers has not been well-studied in ViTs. In this work, we explore low-rank pruning as well as the rank searching on ViT and formulate it as a powerful plug-in.

%To optimizing the self-attention operation, Linformer \cite{DBLP:journals/corr/abs-2006-04768} accelerated the self-attention operation by reducing the rank of the attention matrices and value features based on linear projection. Performer \cite{DBLP:conf/iclr/ChoromanskiLDSG21} approximated the attention matrices in low-rank based on kernelization, and enabled efficient matrix multiplication by rearranging the order. Although these methods work well on Transformer in NLP, in Vision Transformer, linear projection layers occupy the most operation instead of self-attention, as shown in appendix \ref{FLOP analysis}. 
%Targeting linear layer, LRT \cite{DBLP:conf/icassp/WinataCLLF20} utilized the low-rank factorization to approximate the original linear layer with two smaller ones and demonstrate its potential on the speech recognition tasks; however, the effect of low-rank approximation on linear layers is not yet well-studied in Vision Transformer. In this work, we will explore the low-rank pruning on Vision Transformer and dedicate to formulating it as a powerful plug-in.

\subsection{Neural Architecture Search} 
\label{Related:SPOS}

One-Shot NAS~\cite{DBLP:journals/jmlr/ElskenMH19, DBLP:conf/iclr/BrockLRW18, DBLP:conf/iclr/CaiZH19, DBLP:conf/pkdd/StamoulisDWLPLM19, DBLP:conf/eccv/GuoZMHLWS20, liu2018darts} has been proposed to efficiently discover architectures using weight-sharing strategies for CNN. DARTS~\cite{liu2018darts} frames NAS as a constrained
optimization problem and employs differentiable methods to optimize both network parameters and architecture parameters jointly. Interestingly, some prior efforts have adapted DARTS \cite{DBLP:conf/aaai/Xiao0GYSXT023, DBLP:conf/wacv/YuB23} for low-rank architecture search within CNNs. However, such DARTS-based methods often requiring retraining for robust performance, which might be expensive as the number of deployment considerations increase~\cite{DBLP:conf/iclr/CaiGWZH20}.

To surmount the challenge, two-staged based NAS~\cite{DBLP:conf/iclr/CaiGWZH20, DBLP:conf/eccv/GuoZMHLWS20, DBLP:conf/iclr/CaiGWZH20} decouple the training and searching. A supernet is first trained, followed by the application of an evolutionary algorithm to identify the optimal architecture. 

Focusing on the vision transformer, AutoFormer \cite{DBLP:conf/iccv/ChenPFL21} extends the two-staged based NAS paradigm \cite{DBLP:conf/iclr/YuYXYH19}, integrating a weight entanglement strategy to seek the optimal ViT architecture. GLiT \cite{DBLP:conf/iccv/ChenLLL000YO21} further introduces the locality module, incorporating CNN-correlated features into the search space, thus reducing computational costs and explicitly modeling local correlations between patches.

Unlike traditional NAS methods that primarily focus on parameters like kernel size, embedding dimensions, and head numbers in transformers and CNNs, our work centers on the unique low-rank architecture search space. This search space is vast, with each module potentially presenting hundreds of candidate rank settings. Due to the lack of a specifically designed solution for supernet construction, the search process can be burdened by computational overheads and undertraining issues. In this work, we aim to address these challenges. We introduce an optimized One-Shot NAS approach, specially crafted for low-rank architecture selection, to fully harness the potential of low-rank approximation in optimizing ViT.

%In the realm of low-rank architecture selection for transformer, two core challenges prevent the straightforward application of preceding One-Shot NAS techniques: 1) The integration of low-rank transformer subnets, which bear information from pretrained models, into the supernet; 2) Efficient exploration of an expansive search space. To address these, we introduce bi-level Methods, commencing locally and escalating globally, alongside a Low-Rank Aware Supernet crafted for resource efficiency and accelerated convergence. Our methodology stands out, adeptly navigating the intricacies of the low-rank domain, setting it apart from prior NAS approaches.

\begin{figure*}[ht]
    \centering
    \includegraphics[scale=0.47]{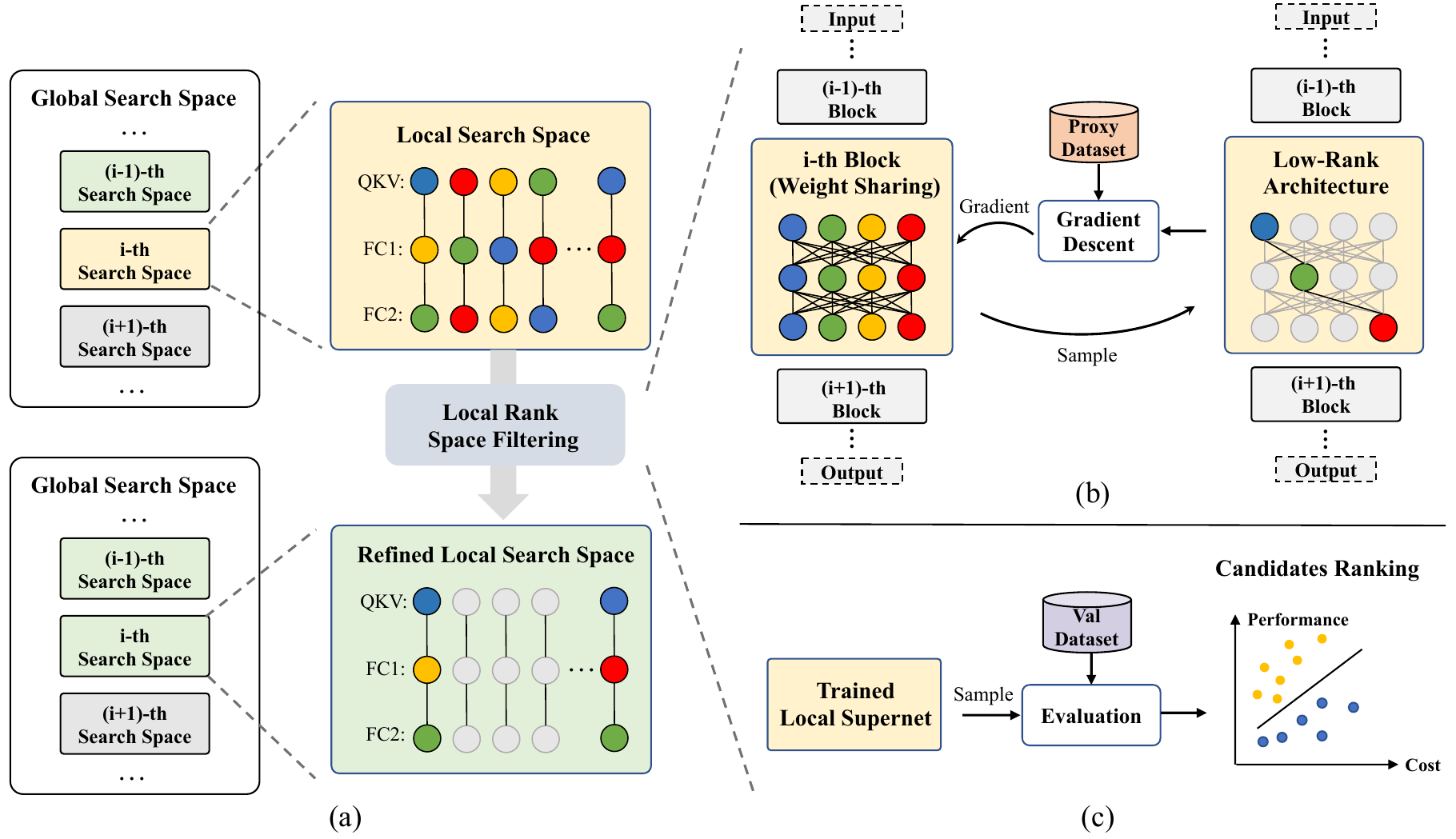}
    \caption{Illustration of low-rank aware candidate filtering. \textbf{(a)}: Filtering is applied sequentially to each block. \textbf{(b)}: For the $i$-th transformer block, we construct a local supernet by substituting the $i$-th block with a weight-sharing superblock that encompasses the local low-rank architecture. \textbf{(c)}: Leveraging the trained supernet as a performance estimator, we filter the local rank space, considering the trade-off between performance benefits and computational costs.}
    \label{fig:/Overview}
\end{figure*}

%\section{Method}
\label{sec:method}
%To formulate the rank selection problem as supernet-based One-Shot NAS paradigm we have to design the three key components: (1). Low-rank search space ; (2) supernet construction and training; 

\section{Preliminary}
\paragraph{One-Shot NAS} In this work, we focus on the two-staged based One-Shot NAS paradigm. First, the architecture search space $\mathcal{A}$ is encoded into an over-parameterized supernet $\mathcal{N}(\mathcal{A}, W)$, where $W$ stands for the weight of supernet and $W$ is shared among candidate architectures (\textit{i.e.}, low-rank architectures $a \in \mathcal{A}$ ). Then, we train the supernet by maximizing the objective function:
\begin{equation}
    \label{SPOS training}
    W_{\mathcal{A}} = \underset{W}{\text{argmin}} \mathbb{E}_{a \thicksim \Gamma(\mathcal{A})} [ \mathcal{L} (\mathcal{N}(a, W(a)) ],
\end{equation}
where $W_{\mathcal{A}}$ is the weights of the supernet, $a \in \mathcal{A}$ is a rank configuration, $W(a)$ is weights of $a$, $\mathcal{N}(a, W(a))$ denotes the corresponding subnetwork, and $\mathcal{L}$ stands for the training loss. 

Second, with the well-trained supernet, we can leverage it as a confident proxy to guide the searching algorithm such as evolutionary algorithm (EA) to search for optimal architecture that maximize the objective function (\textit{e.g.}, accuracy) while statisfying the target constraint (\textit{e.g.}, FLOPs). The objective function can be formulated as: 
\begin{equation}
\begin{aligned}
    \label{SPOS searching}
    & a^* = \underset{a \in \mathcal{A}}{\text{argmax}}\text{ACC}_{\text{val}} (\mathcal{N} (a, W_{\mathcal{A}}(a))) \\
    &\text{s.t.} \quad \mathcal{F}_{lower} \leq \mathcal{F}(a) \leq \mathcal{F}_{upper},
\end{aligned}
\end{equation}
where $\text{ACC}_{\text{val}}$ is the accuracy on validation set, $\mathcal{F}(a)$ denotes the FLOPs of the subnetwork with the configuration $a$, $\mathcal{F}_{lower}$ and $\mathcal{F}_{upper}$ are the lower bound and upper bound of FLOPs constraints, respectively.

%\subsection{Low-Rank Aware Architecture Search}
%\subsection{Low-Rank Search Space} 

%Considering the well-knowed DeiT-S, with 12 blocks and each with 3 compression target, we can reduce the size of the search space from astounding $10^{93}$ to $10^{28}$. 

\section{Methodology}
%\subsection{Problem Formulation}

%In this work, offers an innovative approach to address rank selection by redefining it as a Neural Architecture Search (NAS) problem. This reframing is guided by a key insight: the rank settings determine the architecture of the resulting low-rank model. Furthermore, drawing from the inherent properties of SVD-based low-rank approximation, we propose a solution using the well-established branch known as One Shot NAS. 

%Our objective is twofold. 
\subsection{Mappging Rank Selection into NAS}
\label{subsec:mapping}
Our primary goal is to maximize the benefits of low-rank approximation within transformers. This entails identifying the optimal rank setting \( r \) and then generating the associated low-rank modules for each linear component. In pursuing this, we observed a clear correlation between rank selection and One-Shot NAS, leading to two key insights:

\begin{itemize}

    \item In low-rank approximation, given a rank setting $r$, we decompose the linear modules into two matrices: \( U_r \in \mathbb{R}^{m \times r} \) and \( V_r \in \mathbb{R}^{n \times r} \). Under this scenario, deciding on an optimal rank choice is equivalent to choosing architectural components in network design.   

    \item  The rank-$r$ low-rank approximation inherits the top-most $r$ significant eigenvectors from full-rank pretrained weight matrices. When considering two rank settings $r_i$ and $r_j$, where $r_i < r_j$, it is evident that the set of eigenvectors for $r_i$ is a proper subset of those of $r_j$. During finetuning, this subset property suggests that the low-rank matrices $U_{r_i}$ and $V_{r_i}$ of rank setting $r_i$ can be approximately inherited from $U_{r_j}$ and $V_{r_j}$, respectively. This is because their initial states are identical, with the top $r_i$ eigenvectors holding a significant portion of information from the pretrained weight matrices. This characteristic provides a perfect link for these candidates to be trained jointly, which aligns well with One-Shot NAS and its weight-sharing concept.
    
    %This commonality in eigenvectors provides a perfect link for these candidates to be trained jointly, as the top $r_i$ eigenvectors start in the same state for both, which aligns well with One-Shot NAS and its weight-sharing concept.
     
    %the low-rank counterparts $U_{r_i}$ can be approximately inherited from the $U_{r_j}$. 

        %\item Low-rank approximation inherently preserves the top-most $r$ significant vectors. These influential core vectors remain a sense of consistentency across different rank configurations, facilitating the creation of a genuine weight-sharing supernet with various type of architectural candidates. This characteristic provides a perfect link for these candidates to be trained jointly, which aligns well with One-Shot NAS and its weight-sharing concept.
\end{itemize}

Based on these insights, we propose to map the rank selection problem into NAS and introduce the Fine-grained Low-Rank Architecture Search, abbreviated as \mbox{FLORA}, to effectively identify fine-grained low-rank architecture.

\paragraph{Challenge} 
\vspace{-0.3cm}
While the mapping into One-Shot NAS presents a promising pathway, there are still challenges we must overcome to actualize a sound and effective solution for rank selection. Traditional One-Shot NAS methodologies often focus on a discrete search space with around 3-5 choices per block, which is deemed manageable in terms of computational demand. However, when we consider the realm of rank selection for low-rank approximation, the scenario becomes vastly more intricate, since the candidate rank choices is usually larger than hundreds. 

Earlier studies focused on low-rank approximation for CNNs suggest a simplification of this astounding search space. They advocate for ranks adhering strictly to a constant multiple, often 32 or 64, aiming to strike an optimal balance between search cost and granularity \cite{DBLP:conf/aaai/Xiao0GYSXT023}. However, when these methods are applied to architectures like ViT, the sheer scale of the search space remains overwhelming. For instance, the classic DeiT-B, with its embedding dimension of 768 and a constant multiple of 64, still presents up to 12 rank choices for each linear module.

This expansive choice set introduces two primary challenges. Firstly, without a well-designed strategy to construct and train the supernet, subnetworks may risk undertraining or face weight-coupling issues \cite{DBLP:conf/iccv/Chu0X21, DBLP:journals/corr/abs-1904-00438}. This can lead to biased estimations of network performance. Secondly, once an approximate model has been identified, retraining is often essential to refine its performance – an additional computational overhead that is not desirable, especially in compression scenarios.

To overcome the design challenges mentioned above, we offer a series of techniques to enhance the performance of low-rank supernets, including the filtering strategy to identify weak candidates to prevent from possible interferencing (Sec. \ref{sec:Local Filtering}) and alongside low-rank aware paradigm for supernet consruction and training crafted for resource efficiency and accelerated convergence based on the observation of inherent property of low-rank approximation (Sec. \ref{Sec: Supernet Boosting}).

\subsection{Low-Rank Aware Candidate Filtering}
\label{sec:Local Filtering}
Supernet performance is deeply linked to the quality of its search space. 
When employing low-rank approximation, aggressive rank settings can sometimes lead to irreversible information loss. 
Conversely, architectures with conservative rank settings might not deliver any substantial benefits in terms of computational efficiency. The weight-sharing nature of supernets further complicates this, as tightly coupled subnetwork weights can negatively influence promising architectures, leading to interference and compromised performance.

A seemingly straightforward solution would be to discard these unsuitable configurations. However, the vastness of the low-rank search space renders such a direct approach both cumbersome and computationally taxing. To address this, we draw on a key observation: architectures that underperform at the local level (withing one transformer block) often falter in a global context (entire low-rank architecture). With this in mind, we introduce a bi-level filtering mechanism, called \textit{low-rank aware candidate filtering}, to identify architectures that strike a balance between accuracy and computational demand. The flow of our algorithm is illustrated in Fig. \ref{fig:/Overview}.

\vspace{-0.1cm}
\paragraph{Local Level Filtering} Our strategy commences at the local level, focusing on individual transformer blocks. Technically, we create a local supernet that encompasses the low-rank architecture of the current transformer block, leaving other blocks uncompressed. This local supernet is then trained on a proxy dataset—a subset of the original training data. 
%Impressively, by using just 10\% of the original dataset, this supernet can be trained to serve as an effective estimator in roughly 10\% of the time required for comprehensive global supernet training. 

To steer our selection process, we employ the Precision Cost Ratio, defined as:
\begin{equation}
\mathcal{M}(a) = \lambda * \mathcal{P}(a) - \mathcal{F}(a)
\end{equation}
Here, $\mathcal{P}(a)$ and $\mathcal{F}(a)$ respectively denote the classification accuracy and computational overhead associated with a low-rank architecture $a$. The parameter $\lambda$ adjusts the balance between precision and cost. Utilizing this metric, we rank the low-rank architecture on a block level and retain only the top-$k$ candidates.

\paragraph{Global Level Integration} 
According to the results of local filtering, we then generated the potential global architectures based on the cartesian product. In essence, any global architecture that houses a locally discarded component is eliminated. By doing so, we effectively weed out underperforming candidates, giving rise to a optimized search space, which is smaller and more friendly for supernet construction and training.

\begin{figure}[hpt]
    \centering
    \includegraphics[scale=0.5]{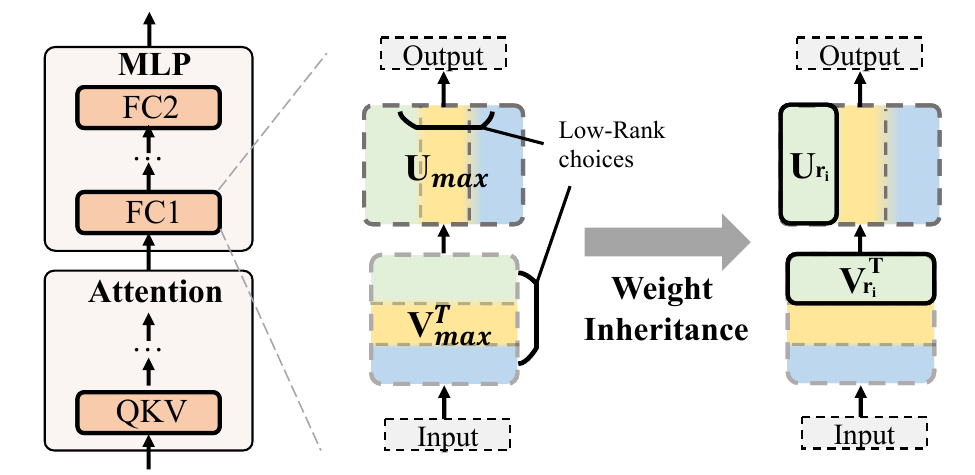}
    \caption{Illustration of weight inheritance technique for supernet construction. All rank architecture of a specific linear modules are encompassed within a superblock. For a designated rank settings, the corresponding rank architecture will be generated by inheriting from superblock}
    \label{fig:LRLinear}
\end{figure}
\vspace{-0.25cm}
\subsection{Boosting Low-Rank Supernet}
\label{Sec: Supernet Boosting}
Achieving a superior supernet performance involves multiple aspects. While refining the search space is crucial, the way subnetworks are incorporated and trained within the supernet is equally important. The successful convergence of the supernet, and its ability to prepare low-rank subnets for immediate use without additional training, depends on both the integration method and the training strategy.
%However, up to our best knowledge, there are not yet a sound strategy for low-rank supernet. 
To address these challenges, we introduce the \textit{Low-Rank Aware Training Paradigm} (LRAT). Tailored for low-rank architectures, LRAT combines two main components: \textit{Weight Inheritance}, which ensures a suitable integration of various rank choices, and \textit{Low-Rank Aware Sampling}, a training approach that adjusts to the needs of different rank architectures. 

\vspace{-0.1cm}
\paragraph{Weight Inheritance} 
For a linear component with $N$ candidate rank choices $\mathbb{C}=\{r_1, r_2, ... , r_N\}$ and we can yield corresponding low-rank module set $\mathbb{S}=\{(U_{r_1}, V_{r_1}), (U_{r_2}, V_{r_2}) , ... , (U_{r_N},  V_{r_N})\}$. Our goal is to design a suitable technique to integrate all of these candidates into a choice block so that our supernet can be constructed.

In our earlier discussion (Sec. \ref{subsec:mapping}), we explored how top eigenvectors overlap across different rank choices in low-rank approximations. Specifically, for a rank $r_i$, the top eigenvectors, originating from the full-rank pretrained weight matrices, capture a significant chunk of information from the initial state. Furthermore, these eigenvectors for $r_i$ act as a subset to the eigenvectors of a higher rank $r_j$ where $r_i < r_j$. This suggests that as we navigate through different rank settings, there is both reusability and overlap of information from the initial state across different ranks. Consequently, instead of handling each low-rank module in isolation, we could co-train the corresponding modules by jointly sharing their parameters and gradients to boost the convergence. 

Building on these insights and drawing inspiration from the shared convolutional kernels in CNN NAS practices \cite{DBLP:conf/pkdd/StamoulisDWLPLM19, DBLP:conf/iclr/YuYXYH19}, we present the \mbox{\textit{weight inheritance}} method. As illustrated in Fig. \ref{fig:LRLinear}, every low-rank module is encapsulated within the largest one, expressed as:
\begin{equation}
\label{Eq:Weight inheritance1}
U_{r_i} = U_{r_N}[:, : r_i],
\quad V_{r_i} = V_{r_N}[:, : r_i]
\end{equation}
For a specific rank $r_i$, the values of $U_{r_i}$ and $V_{r_i}$ are inherited from the top-$r_i$ columns of $U_{r_N}$ and $V_{r_N}$ respectively. During the backward pass, the gradients of $U_{r_i}$ and $V_{r_i}$ are updated back to the corresponding sub-matrices in $U_{r_N}$ and $V_{r_N}$. Within each choice block, the candidate with the largest rank is set to the size of the uncompressed model's factorized weights and with SVD.

It's evident that each sub-structure is a subset of the structures with higher rank levels, culminating in the super-matrix, as shown below: 
\begin{equation}
\label{Eq:Weight inheritance2}
\begin{gathered}
    U_{r_i} \subseteq U_{r_k} ,\quad V_{r_i} \subseteq V_{r_k} \\
    \forall k \in\{i+1, i+2, ..., N\} \quad \forall r_i \in \mathbb{C}
\end{gathered}
\end{equation}
With shared weights among low-rank modules, gradient information can be efficiently utilized across groups. This mutual gradient update strategy ensures that all candidate low-rank architectures are trained concurrently, bolstering convergence during the supernet training phase.

\paragraph{Low-Rank Aware Sampling}
Contrary to prior works \cite{DBLP:conf/eccv/GuoZMHLWS20, DBLP:conf/iccv/ChenPFL21} that train the supernet from scratch, our approach initializes the supernet with weights from a pretrained uncompressed model. In this context, subnetworks are seen as low-rank approximated networks for different rank choices. This introduces a new perspective on sampling during training: it's suboptimal to sample subnetworks using a uniform distribution. Instead, our design for a new sampling distribution is guided by two key observations:
\begin{itemize}
    \item Architectural search algorithms are generally complex and hard to train. For a pre-selected search space, it's computationally arduous to thoroughly train every substructure. For the entire system to achieve peak performance and consistent convergence, training resources must be judiciously allocated to different subnetworks, warranting a non-uniform distribution.
    \item Low-rank architectures, especially those of lower ranks, derive varying degrees of information from pretrained weights. Such inheritance can be perceived as the architecture having undergone preliminary training. As a result, during supernet training, it's crucial to divert more resources (like sampling probability) towards these less-informed structures, ensuring balance.
\end{itemize}
With the above insights in mind, we introduce a non-uniform path sampling strategy that favors smaller rank choices within each low-rank linear layer.
Formally, let's define the rank choice for a low-rank linear layer by a random variable $X$. 
Its Probability Mass Function (PMF) of $X$ is formulated as:
\begin{equation}
\label{Eq:NUS}
\text{$ p_X(r) = P(X=r) = \frac{\frac{1}{r}}{\sum_{r'\in \mathbb{C}}\frac{1}{r'}} ,\ \ \  \ \ r \in \mathbb{C} $}
\end{equation}
Recall that $\mathbb{C}$ denotes the rank choice set of a specific low-rank linear layer. The prior distribution of selecting a sequence of rank choice would be $\Gamma(\mathcal{A}) = P(X_1 = r_{k_1}, ..., X_i = r_{k_i} , ..., X_l = r_{k_l})$, where $X_i, r_{k_i}$ denotes the random variable of $i$-th choice block and its correlated rank choice, respectively.

\vspace{-0.2cm}
\section{Experiments}
\label{sec:experiment}

%The proposed methodology is denoted as \textbf{FLORA} by FLORAming the transformer structure through searching for a low-rank approximation.
We trained and tested FLORA on the ImageNet-1k \cite{DBLP:conf/cvpr/DengDSLL009} dataset using several representative ViT models, such as DeiT \cite{DBLP:conf/icml/TouvronCDMSJ21} and Swin Transformer \cite{DBLP:conf/iccv/LiuL00W0LG21}, as our pre-trained backbones. In line with earlier NAS practices \cite{DBLP:conf/eccv/YuJLBKTHSPL20, DBLP:conf/iccv/YuH19}, our supernet is trained under the in-place distillation strategy (e.g., under the supervision of the uncompressed model itself). We employed the evolutionary algorithm to search for the target architecture. For detailed information on hyperparameters used in training and supernet configuration, please refer to the supplemental materials. 

%We conduct all experiments with 8 NVIDIA V100 GPUs using Pytorch \cite{NEURIPS2019_9015} framework and timm library \cite{rw2019timm}. 

%For the choice of the pre-trained teacher, we conduct two sets of experiments with different types of teachers for distillation. 

% \shixing{A quick question, is the soft-label group all conduct distillation our uncompressed models? and the strong distillation group all conduct distillation on strong teachers?}
% \chichih{Yes, we group the result using our uncompressed models at sec 4.1, and strong teacher in sec 4.2, and both these two groups use the soft-label only distillation introduce in methodology.}
% \shixing{I don't quite understand the settings right here, maybe we can meet sometimes to discuss.}

%First, for a fair comparison and to demonstrate the efficacy of our framework in restricted situations where a strong teacher is unavailable, we select the uncompressed model itself for the supervising task. Second, to unleash the potential of the low-rank approximation, we further resort to the larger-scale teacher model Swin-L \cite{DBLP:conf/iccv/LiuL00W0LG21}, that is trained on a larger dataset so that the student can not only restore lost information but also distill some extra knowledge from the teacher. In the following section, we denote the soft-label distillation training with and without a large-scale model as \emph{self-distillation} and \emph{strong-distillation}, respectively.

\begin{table}[h]
\centering
\caption{Comparison of FLORA with different ViT compression methods on ImageNet-1k dataset.}
\resizebox{7.5cm}{!}{
  \begin{tabular}{c  r c c c c c }
  \toprule
   Method & FLOPs & \begin{tabular}{@{}c@{}}FLOPs \\ saving \end{tabular} &  Params & \begin{tabular}{@{}c@{}}Params\\ saving \end{tabular}  &\begin{tabular}{@{}c@{}}Top-1\\ Acc \end{tabular}  \\
   \midrule
    \multicolumn{6}{c}{\textbf{DeiT-S}} \\
    \midrule
    Baseline    & 4.7G &  -     & 22.1M & -     & 79.8\%  \\
    DynamicViT  & 3.4G & 28\%   & 23.1M & -     & 79.6\% \\
    SPViT       & 3.3G & 30\%   & 16.4M & 26\%  & 78.3\% \\
    WDPruning   & 3.1G & 32\%   & 15.0M & 32\%  & 78.6\% \\
    S$^2$ViTE   & 3.1G & 32\%   & 14.6M & 34\%  & 79.2\% \\
    %DynamicViT & 2.9G & 38\% & 79.3\% & 0.5\% \\
    % &   MDC         & 2.9G & 38\%   & -     & -     & 79.9\% &  0.1\% \\
    UVC         & 2.7G & 42\%   & -     & -     & 79.4\%  \\

   \rowcolor{MyPink}\textbf{Ours} & 2.7G & 42\% & 12.6M & 43\% & 79.6\% \\
   \midrule
   \multicolumn{6}{c}{\textbf{DeiT-B}} \\
    \midrule
   Baseline & 17.6G & -  & 86.4M & -& 81.8\% \\
   %&VTP & 13.8G & 22\% & 67.3M & 22\% &81.3\% & -0.5\% \\
   S$^2$ViTE       & 11.8G & 33\% & 56.8M  & 35\%  & 82.2\%        \\
   SPViT           & 11.7G & 33\% & 62.3M  & 28\%  & 81.6\%       \\
   DynamicViT      & 11.2G & 36\% & 87.2M  & -     & 81.3\%        \\
   WDPruning       & 11.0G & 37\% & 60.6M  & 30\%  & 81.1\%      \\
   %&PatchSlimming   & 9.8G  & 44\% & -      & -     & 81.5\%        & -0.3\% \\
   UVC             & 8.0G  & 55\% & -      & -     & 80.6\%        \\
   % &MDC             & 7.0G  & 60\% & -      & -     & 81.5\%        & -0.3\% \\
   %\rowcolor{MyPink} 
   %\textbf{Ours} & 11.1G & 36\% & 53.6M & 38\% & 82.2\% \\
   
   \rowcolor{MyPink} 
   \textbf{Ours} & 8.0G & 55\% & 37.9M & 56\%  & 81.8\%  \\
 \midrule
   \multicolumn{6}{c}{\textbf{Swin-S}} \\
    \midrule
    Baseline & 8.7G & - & 49.6M & -& 83.2\% \\
    DynamicViT & 6.9G & 20\% & 50.8M & - & 83.2\%  \\
    WDPruning & 6.8G & 22\% & 37.4M & 26\% &82.4\% \\
    SPViT  & 6.1G & 30\% & 39.2M & 30\% & 82.4\% \\
    %  \rowcolor{MyPink} 
    %\textbf{Ours} & 6.6G & 25\% &  37.3M & 26\% & 83.2\% \\
    \rowcolor{MyPink} 
    \textbf{Ours} & 6.1G & 30\% & 34.8M & 30\% & 82.8\% \\
    \midrule
    \multicolumn{6}{c}{\textbf{Swin-B}} \\
    \midrule
    Baseline & 15.4G & - & 87.8M & - & 83.5\%  \\
    DynamicViT & 12.1G & 21\% & 88.8M  & -  & 83.4\% \\
    SPViT  & 11.4G & 26\% & 68.0M & 24\% &83.2\% \\
    \rowcolor{MyPink} 
   %\textbf{Ours} & 11.4G & 26\% & 63.5M & 28\%  & 83.7\%\\
   \textbf{Ours} & 9.0G & 42\% & 52.3M & 41\%  & 83.2\%\\
   % \midrule
   %\multicolumn{5}{l}{\emph{\textbf{AutoFormer-B}}} \\
   %Baseline & 11.3G & -  & 82.4\% & -\\
   %\rowcolor{MyPink} 
   %\textbf{Ours} & 7.5G & 34\% & 82.7\% & +0.3\% \\
   %\midrule
   %\multicolumn{5}{l}{\emph{\textbf{AutoFormer-S}}} \\
   %Baseline & 5.1G & -  & 81.7\% & -\\
   %\rowcolor{MyPink} 
   %\textbf{Ours} & 3.6G & 30\% & 81.4\% & -0.3\% \\
   %\midrule
   %\multicolumn{5}{l}{\emph{\textbf{AutoFormer-T}}} \\    Baseline & 1.3G & -  & 74.7\% & -\\
   % \rowcolor{MyPink} 
    %\textbf{Ours} & 1.2G & 10\% & 74.9\% & +0.2\% \\
    \bottomrule
    \end{tabular}
    }
  \label{Comparison Table}
\end{table}

\subsection{General Comparison}
%For the choices of the teacher model for soft-label distillation, we first select the uncompressed model itself for supervising for fai.
We compare our results with state-of-the-art ViT pruning methods, ranging from input sequence reduction (DynamicViT \cite{DBLP:conf/nips/RaoZLLZH21}), weight pruning (WDPruning \cite{DBLP:conf/aaai/0004HWCCC22}, S$^2$ViTE \cite{DBLP:conf/nips/ChenCGYZW21}), and multi-dimension pruning (UVC \cite{DBLP:conf/iclr/YuCSYTY0W22}). It is worth mentioning that DynamicViT and UVC both include the knowledge distillation in their methodologies, which is identical to our method.

%MDC \cite{DBLP:conf/cvpr/HouK22a}). 
%Here, all candidates used in the comparison use no extra data and rely on strong teachers for knowledge distillation. 
The results are shown in Tab. \ref{Comparison Table}. 
Compared to channel pruning methodologies (WDPruning~\cite{DBLP:conf/aaai/0004HWCCC22}), we achieve around 1\% accuracy improvement under the same or smaller level of computation reduction and model size on every backbone, demonstrating the effectiveness of searching for a low-rank subnetwork against structure pruning techniques.
Besides, when considering the sequence reduction-based methods (DynamicViT~\cite{DBLP:conf/nips/RaoZLLZH21}), FLORA achieves comparable performance under with better FLOPs savings with different optimized targets. 
For \mbox{DeiT-B}/\mbox{Swin-B}, we can either enjoy 20\%/21\% additional FLOPs reduction with competitable or even higher top-1 accuracy. Besides, FLORA further demonstrates its superiority with not only FLOPs saving but an extra 60\%/30\% model size reduction when compared to DynamicViT. Lastly, when compared to multi-dimensional compression methods, we achieve superior results on DeiT-Base, registering a 1.2\% improvement in performance under the same FLOPs level. Overall, we can observe that FLORA shows competitive performance with SOTA methods on small models like DeiT-S and Swin-S while outperforming all the methods under comparison on large models like DeiT-B and \mbox{Swin-B}.

\begin{figure}[ht]
    \centering
    \includegraphics[scale=0.28]{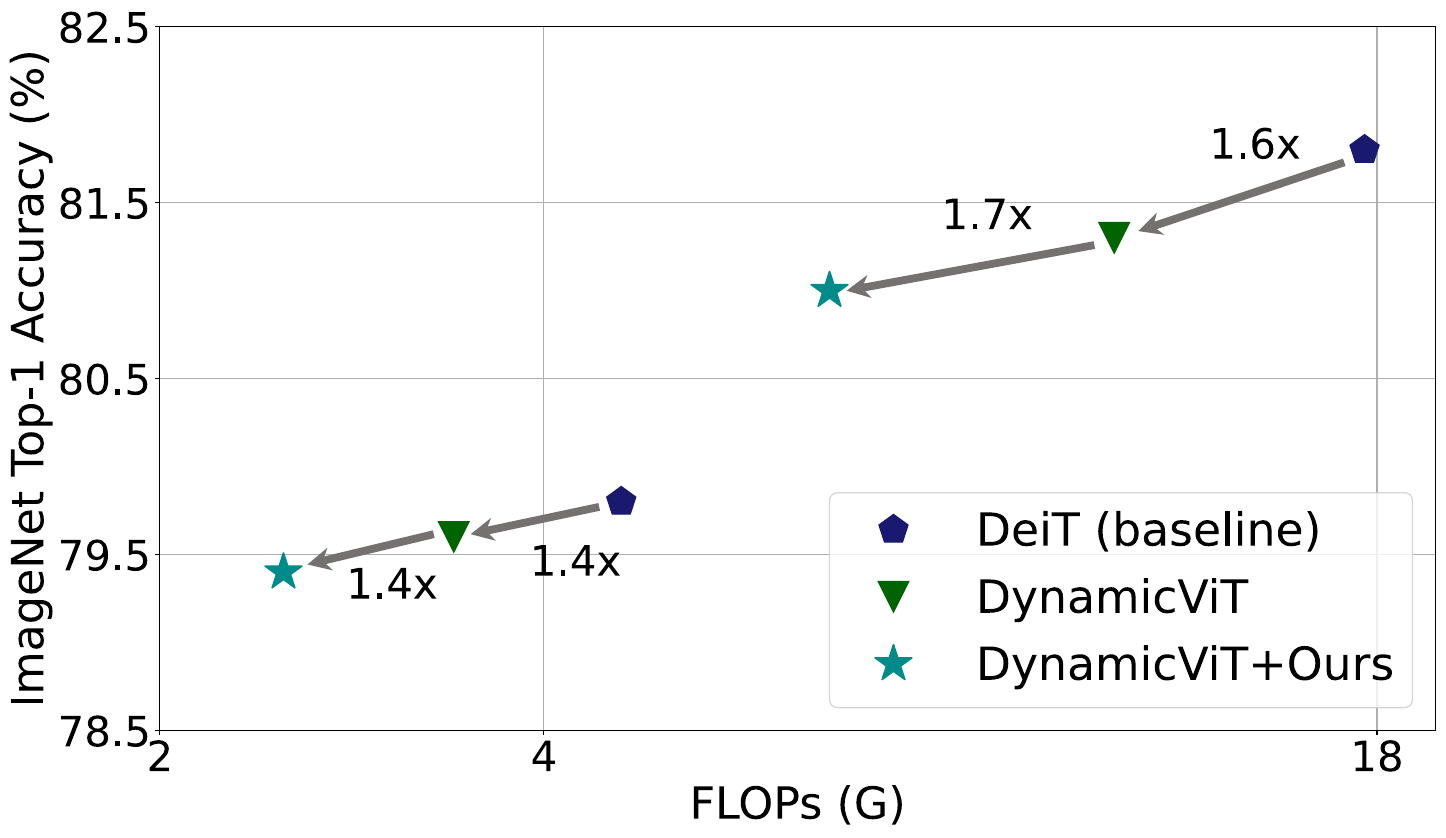}
    \caption{Results of integrating FLORA with DynamicViT.}
    \label{Orthogonality}
\end{figure}

\begin{table}[]
\centering
\caption{Experiments on TinyViT and integration with DynamicViT. DynamicViT is denoted as DyViT. Results show the orthogonality of FLORA on top of other compression methods }
\label{GeneralandOrtho}
\resizebox{6.8cm}{!}{
    \begin{tabular}[b]{c | l | l l}
    \toprule
    Model & Methods & GFLOPs & Top-1 Acc \\
    \midrule
    \multirow{2}{*}{TinyViT-21M}  & baseline & 4.3G & 83.1\% \\
    & \textbf{Ours} & \textbf{3.2G \textcolor{blue}{(-25\%)}} & \textbf{83.1}\% \textbf{\textcolor{blue}{(-0\%)}} \\ 
    %\midrule
    %TinyViT-21M-22k & 4.3G & 84.8\% \\
    %FLORA-TinyViT-21M-22k & 3.9G \textcolor{blue}{(-10\%)} &84.8\% \textcolor{blue}{(-0\%)} \\ 
    \midrule 
    \multirow{3}{*}{DeiT-S} & baseline &4.7G & 79.8\%  \\
    & DyViT &3.4G \textcolor{blue}{(-28\%)}& 79.6\% \textcolor{blue}{(-0.3\%)} \\
    & \textbf{DyViT + Ours} & \textbf{2.3G} \textbf{\textcolor{blue}{(-49\%)}} & \textbf{79.3\% \textcolor{blue}{(-0.6\%)}} \\
    \midrule
    \multirow{3}{*}{DeiT-B}  & baseline          & 17.6G & 81.8\%  \\
     & DyViT     & 11.2G \textcolor{blue}{(-36\%)}& 81.3\% \textcolor{blue}{(-0.5\%)} \\
     & \textbf{DyViT + Ours} & \textbf{6.70G} \textbf{\textcolor{blue}{(-62\%)}}& \textbf{81.0\%} \textbf{\textcolor{blue}{(-0.8\%)}} \\
    \bottomrule
    \end{tabular}
    
}
   \vspace{-0.2cm}
\end{table}
\vspace{-0.2cm}
\subsection{Generality and Orthogonality}
%\textbf{Throughputs}.
\vspace{-0.1cm}
\paragraph{Results on the Compact Hybrid Transformer}
In the previous section, we have demonstrated the generality of FLORA on Transformer-only variants DeiT and Swin Transformer. 
Here we further show that our approach is agnostic to model architectures even on a hybrid structure (\emph{i.e.}, CNN + Transformer).
The results of FLORA a TinyViT-22M \cite{DBLP:journals/corr/abs-2207-10666} with 224 x 224 input is shown in Tab. \ref{GeneralandOrtho}. 
The computational cost can be reduced by 25\% without sacrificing performance on TinyViT, affirming the generality of our proposed FLORA.

% \shixing{I think the result on FLORA-TinyViT-21M-22k is quite minor, is there other hybrid architecture(e.g. LeViT) that we can try for orthogonality? If not, I would suggest we do not keep this data and run on some other architectures for rebuttal}

\vspace{-0.4cm}
\paragraph{Integrating with Other Compression Methods}
To show the orthogonality of our approach, we integrate the proposed FLORA with token reduction based compression method DynamicViT to achieve further compression ratio. Here, we use the uncompressed model itself for supervising. From Tab. \ref{GeneralandOrtho}, FLORA achieves 21\%/26\% more FLOPs reduction while only sacrificing 0.3\% more performance degradation on DeiT-S/B. 
Fig. \ref{Orthogonality} provides an clearer view for the demonstration of orthogonality. 
On DeiT-B, when integrating FLORA with DynamicViT, the reduction on FLOPS is further improved by 1.7x over standalone DynamicViT. 
Moreover, on the small-scale model DeiT-S, we can also earn 1.4x more FLOPs improvement and a compelling 50\% FLOPs savings in total. 
Thus, FLORA provides a new perspective that is orthogonal to the previous compression methods. 
By applying multiple compression techniques, ViT can be compressed to an appealing ratio with less than 50\% of the original FLOPs and less than 1\% accuracy drop.

\iffalse
\paragraph{Integrating with Other Neural Architecture Search Methods}
Aside from the compression techniques mentioned above, our framework can also be generally applied to the optimized network searched by the previous NAS method. As shown in Tab. \ref{GeneralandOrtho}, on top of the representative One-Shot NAS methods~\cite{DBLP:conf/iccv/ChenPFL21}, which search the number of heads, mlp ratio, number of layers, and embedding dimensions, the proposed FLORA can further improve the FLOPs by up to 34\% with negligible performance degradation. 
% For fairness comparison, we align the distillation setting to AutoFormer using RegNetY\cite{radosavovic2020designing} as the teacher and considered the hard label together while doing distillation.
The results again evidences the remarkable potential and generality of the proposed FLORA to further push the compression ratio higher by exploiting redundancy from the search space of rank. 
\fi
%Besides, as mentioned in Sec \ref{Related_work:}, the low-rank approximation represents an independent direction of optimization to the other aspects that other works focus on; thus, our framework can not only be packed together with these methodologies but also serve as a new pruning candidate that multi-dimensional compression method can take into consideration.

\begin{figure}[htp]
    \centering
    \includegraphics[width=.45\textwidth]{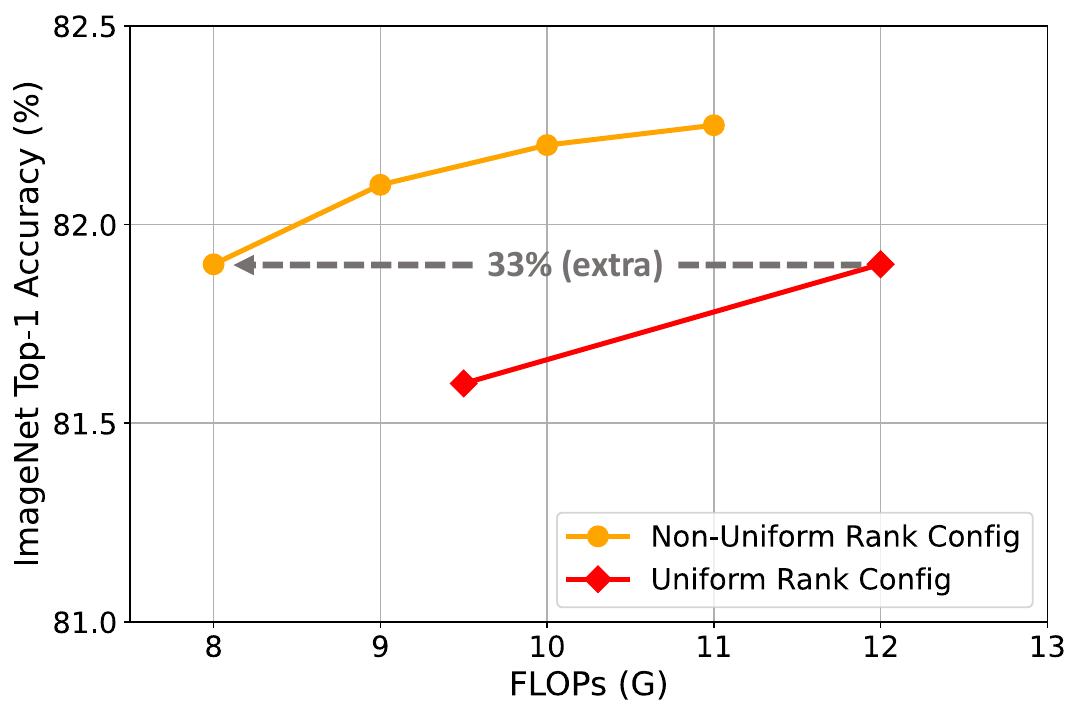}
    \caption{Performance difference of using different rank config for Low-rank approximation on DeiT-B.}
    \captionsetup{justification=centering,margin=2cm}
    \label{Uniform v.s Non-Uniform}
\end{figure}

\subsection{Benefit of Non-Uniform Rank Setting}
In this section, we study the impact of employing fine-grained, non-uniform rank configurations on the model's performance. We utilize DeiT-B as the backbone and proceed with training a low-rank supernet. To ensure fair comparison, all considered low-rank architectures are derived from this trained supernet. As illustrated in Fig. \ref{Uniform v.s Non-Uniform}, the advantages of a non-uniform rank configuration become apparent. When comparing under the same accuracy level, a non-uniform rank configuration can achieve up to 33\% additional FLOPs reduction relative to its uniform counterpart \cite{DBLP:conf/icassp/WinataCLLF20}. These findings underscore the value of layer-wise non-uniform rank configurations in the low-rank approximation for ViT and further emphasize to the proficiency of FLORA in uncovering these configurations.

%Following the early practices for transformer \cite{DBLP:conf/icassp/WinataCLLF20}, we use the LRA with a uniform rank configuration and finetune the compressed model with the same epochs to generate the baseline for a fair comparison. From Fig. \ref{Uniform v.s Non-Uniform}, we can see the benefit of the non-uniform rank configuration. Under the same accuracy level, a searched non-uniform rank configuration can yield up to 25\% extra FLOPs reduction. 

\begin{figure}[ht]
    \centering
    \includegraphics[scale=0.2]{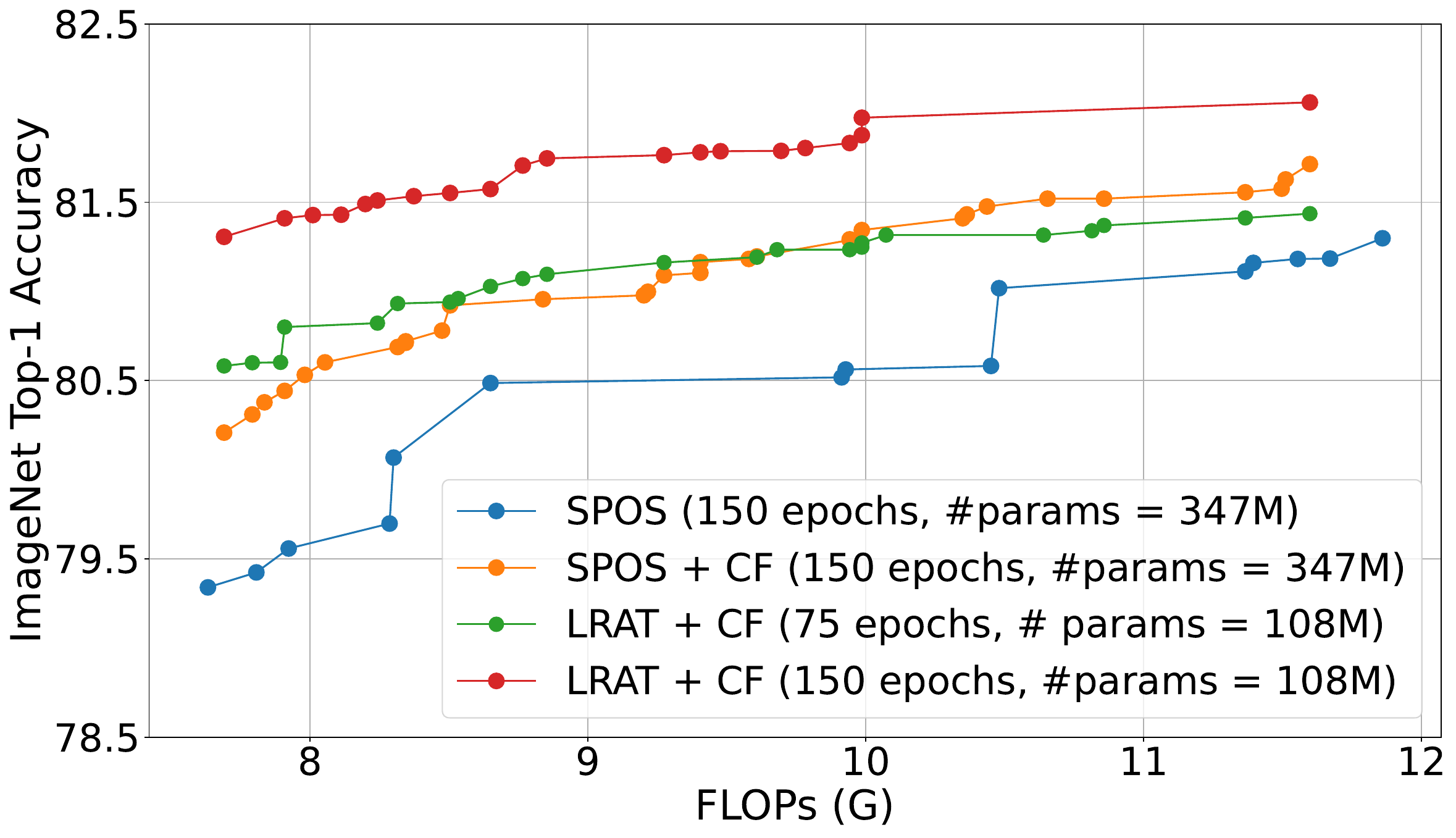}
    \caption{Pareto frontier of subnetworks sampled from low-rank supernet of DeiT-B with different strategies. \#params denote the number of trainable parameters of the supernet. CF: Low-Rank Aware Candidate Filtering, LRAT: Low Rank Aware Supernet Training Paradigm}
    \label{fig:ablation study}
\end{figure}

\iffalse
\begin{table}[]
\centering
\caption{Performance comparison of searched low-rank subnets of Deit-B via different paradigm.}
\label{tab:Searched Comp}
\resizebox{6cm}{!}{
    \begin{tabular}[b]{c | c c c}
    \toprule
    Methods & Params & GFLOPs & Top-1 Acc \\
    \midrule
    SPOS & 37.9M & 8G & 80.2\% \\
    \midrule
    Bi-level (Ours) & 37.9M & 8G & 81.1\% \\
    \bottomrule
    \end{tabular}
}
   \vspace{-0.2cm}
\end{table}
\fi
\vspace{-0.2cm}
\subsection{Ablation Study}

\label{Sec:Analysis}
In this section, we conduct an ablation study to analyze the impact of our \textit{Low-Rank Aware Candidate Filtering} and the proposed \textit{Low-Rank Aware Training Paradigm} on enhancing supernet quality. The results are visualized using Pareto frontier analysis in Fig. \ref{fig:ablation study}. Our experiments are based on the DeiT-B backbone. As a baseline, we employ the well-established One-Shot NAS method, SPOS\cite{DBLP:conf/eccv/GuoZMHLWS20}, which was originally designed for CNN search spaces. We have adapted SPOS to accommodate the low-rank search space, while preserving its original settings. In the course of supernet construction, each low-rank module within the choice blocks is treated as an independent entity. We train the SPOS supernet under the same hyperparameters as ours. During the supernet's training phase, architectures are selected through uniform sampling. 
\vspace{-0.2cm}
\paragraph{Efficacy of Low-Rank Aware Candidate Filtering} As elaborated in Section Sec. \ref{sec:Local Filtering}, our low-Rank aware candidate filtering focuses on pinpointing weaker architectural components and then uses this insight to generate the refined search space.
The marked benefits of our filtering technique on supernet performance are illustrated in Fig. \ref{fig:ablation study}. Using the same training approach as the SPOS baseline, the performance of subnetworks across diverse FLOP levels sees a notable boost with the incorporation of our filtering approach. Notably, for subnetworks operating around the 8G FLOP mark, we observe a nearly 1\% enhancement in Top-1 accuracy. This considerable enhancement underscores the proficiency of our low-rank aware candidate filtering in pinpointing and eliminating subpar architectures, effectively reducing the training interference from such suboptimal configurations.

\vspace{-0.2cm}
\paragraph{Efficacy of Low-Rank Aware Training Paradigm} Our approach to low-rank aware supernet training combines the \textit{Weight Inheritance} technique with \textit{Low-Rank Aware Sampling}, both specifically designed for low-rank scenarios. Notably, our supernet, trained for only 75 epochs using this approach, performs as well as the SPOS paradigm which requires 150 epochs, emphasizing a faster convergence. When trained for the full duration, our results sit on the Pareto frontier, demonstrating the efficacy of our paradigm.

\vspace{-0.2cm}
\section{Conclusion}
\label{sec:conclusion}
\vspace{-0.1cm}
In this work, we focus on the rank selection problem, exploring the integration of low-rank approximation into the Vision Transformer (ViT) architecture. We observe a notable similarity and alignment between the processes of rank selection and One-Shot NAS. Motivated by this observation, we introduce FLORA—an end-to-end framework based on NAS—to autonomously search for the fine-grained low-rank configurations. To bolster the effectiveness of this approach, we also introduce a series of specialized techniques tailored to enhance the quality of the low-rank supernet. Extensive experiments from our research demonstrates the potential of low-rank approximation as a effective compression technique for the optimization of transformers. For instance, on DeiT-B, with fine-grained rank configuration up to 55\% FLOPs can be saved without performance drop. This discovery holds immense implications, particularly considering the current void in methods that automate rank selection for ViT.

%We believe FLORA aptly addresses this gap, offering a promising tool for the ViT community.

 %Extensive experiments show strong performance and good generality. Extending our framework for multi-dimensional compression can be an exciting direction in the future. 

%\section*{Acknowledgments} 
%We thank to National Center for High-performance Computing (NCHC) of National Applied Research Laboratories (NARLabs) in Taiwan for providing computational and storage resources.

\section*{Acknowledgements}
This work was supported in part by NSF CCF Grant No. 2107085, ONR Minerva program, and iMAGiNE — the Intelligent Machine Engineering Consortium at UT Austin. We thank to National Center for High-performance Computing (NCHC) of National Applied Research Laboratories (NARLabs) in Taiwan for providing computational and storage resources.

%%%%%%%%% REFERENCES
{\small
\bibliographystyle{ieee_fullname}
\bibliography{egbib}
}

\end{document}